\documentclass[10pt,twocolumn,letterpaper]{article}

\usepackage{cvpr}
\usepackage{times}
\usepackage{epsfig}
\usepackage{graphicx}
\usepackage{amsmath}
\usepackage{amssymb}
\usepackage{multirow}
\usepackage{booktabs}
\usepackage{indentfirst}

\usepackage[breaklinks=true,bookmarks=false]{hyperref}
\cvprfinalcopy % *** Uncomment this line for the final submission

\def\cvprPaperID{****} % *** Enter the CVPR Paper ID here

\ifcvprfinal\fi
\begin{document}

%%%%%%%%% TITLE
%title{\LaTeX\ Author Guidelines for CVPR Proceedings}
\title{Correlating Edge, Pose with Parsing}

%\quad 
\author{Ziwei Zhang$^1$,  Chi Su$^{2*}$,  Liang Zheng$^3$,  Xiaodong Xie$^{1}$\\
$^1$School of Electronics Engineering and Computer Science, Peking University\\ $^2$Kingsoft Cloud,  $^3$Australian National University \\
\tt\small {\{ziwei.zh,donxie\}@pku.edu.cn}, \tt\small  {suchi@kingsoft.com}, \tt\small  {liang.zheng@anu.edu.au}
}

%\and
%Chi Su\\
%Kingsoft\\
%{\tt\small suchi@kingsoft.com}
%}

\maketitle
\thispagestyle{empty}

%%%%%%%%% ABSTRACT
\begin{abstract}
According to existing studies, human body edge and pose %when singly deployed, 
are two beneficial factors to human parsing. The effectiveness of each of the high-level features (edge and pose) is confirmed through the concatenation of their features with the parsing features. Driven by the insights, this paper studies how human semantic boundaries and keypoint locations can jointly improve human parsing. Compared with the existing practice of feature concatenation, we find that uncovering the correlation among the three factors is a superior way of leveraging the pivotal contextual cues provided by edges and poses. To capture such correlations, we propose a Correlation Parsing Machine (CorrPM) employing a heterogeneous non-local block to discover the spatial affinity among feature maps from the edge, pose and parsing. The proposed CorrPM allows us to report new state-of-the-art accuracy on three human parsing datasets. Importantly, comparative studies confirm the advantages of feature correlation over the concatenation. %Code is available at: {\color{magenta}{\small https://github.com/ziwei-zh/CorrPM}.}
\end{abstract}

\let\thefootnote\relax\footnotetext{*Corresponding Author.}
\let\thefootnote\relax\footnotetext{Code is available at: \small https://github.com/ziwei-zh/CorrPM.}

%%%%%%%%% BODY TEXT
\section{Introduction}\label{section:intro}
This paper studies human parsing, aiming to partition a human image into semantic regions including body parts and clothes. This problem is challenging due to the complicated textures and styles of clothes, the deformable human body, the scale diversity of different categories, \emph{etc}. As such, directly applying general semantic segmentation methods to human parsing may lead to unsatisfying results, which are reflected in two aspects. First, the boundaries between adjacent parts may be inaccurately located. %it causes \textbf{boundary localization inaccuracy problem} between two parts. 
The system might get confused with pixels along the boundaries, 
especially when the neighboring parts have similar appearance. Second, semantics of segmented  parts may be inconsistent with human body structure, if we don not consider the affinity among different parts. This leads to mislabeling or missing predictions when context clues are not obvious.
\begin{figure}
	\begin{center}
		%\fbox{\rule{0pt}{2in} }
		\rule{.9\linewidth}{0pt}			\includegraphics[width=1\linewidth]{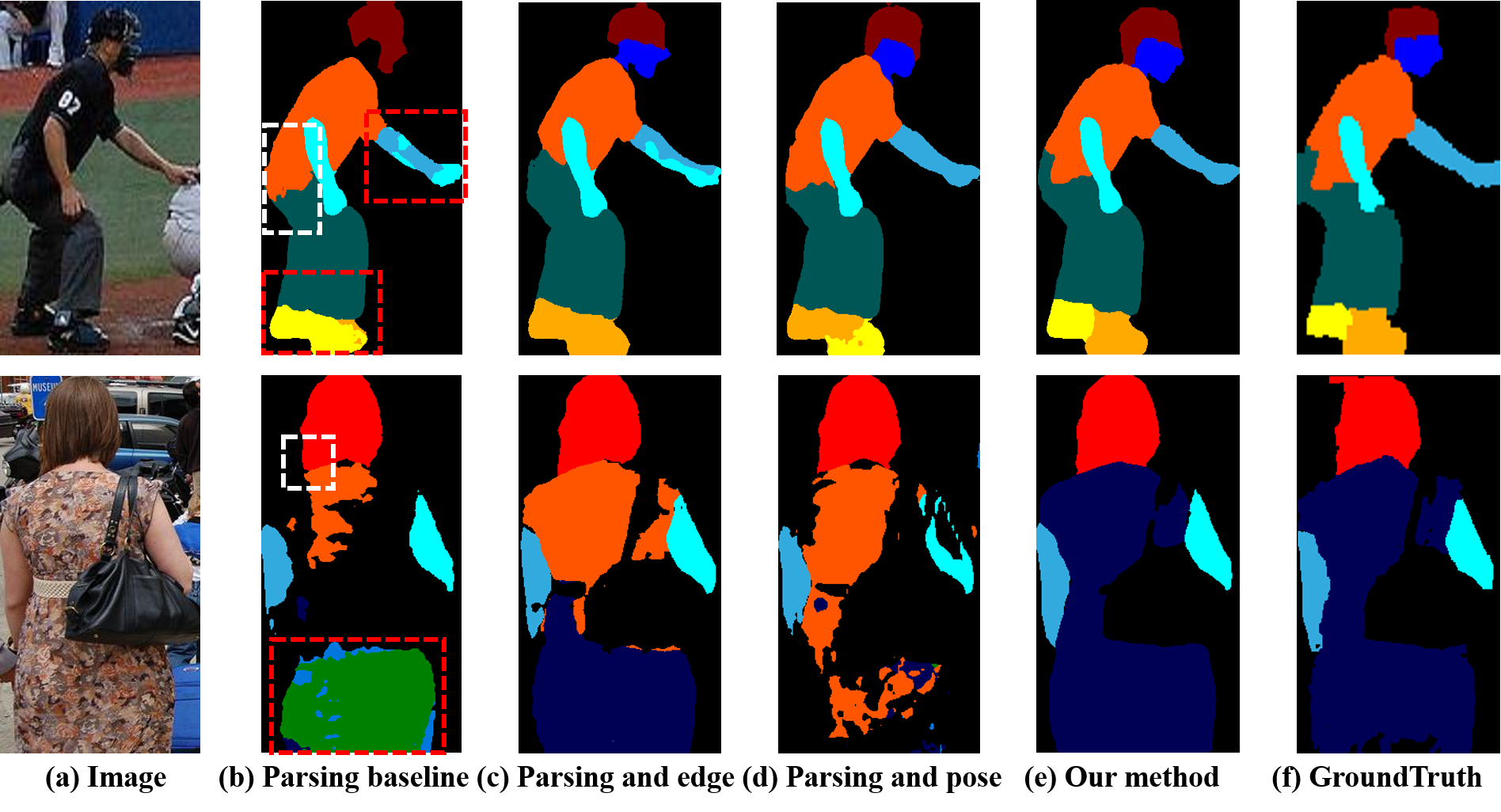}
	\end{center}
	\vspace{-3mm}
	\caption{Illustration of parsing errors and our motivation. (a) Given images. (b) Results of parsing baseline \cite{Chen2017deeplab}. (c) Fusion of parsing and human body edge features. (d) Fusion of parsing and the human keypoint features. (e) Results of our method. (f) Groundtruth. 
	From (b), we observe parsing errors happen due to boundary ambiguity (white box) and body structure inconsistency (red box), respectively. The fusion of boundary features (c) or keypoint features (d) may mitigate one of the two errors.  %is mitigated The at the part boundaries and  and some existing methods fuse parsing features with either (c) human boundary features or (d) human keypoint features. 
	The two types of errors are obviously mitigated in (e) because we take the advantage of both boundary and keypoints by learning their correlation with parsing.
	By comparison, the proposed strategy is superior to concatenation or post processing as commonly done. 
	}
	\label{fig:motivation}
	\vspace{-3mm}
\end{figure}
 
Edge detection and pose estimation can potentially address the above two problems. 
For the first problem, \emph{i.e.,} boundary confusions, human body edge detection is beneficial to distinguish two adjacent categories \cite{chen2016semantic,ruan2019devil}. For the second problem, \emph{i.e.,} semantics inconsistency, pose estimation provides keypoints to enforce the parsing results to be semantically consistent with human body structure  \cite{xiao2018simple,newell2016stacked,sun2019deep}. %can provide human keypoint cues which help the parsing results to be semantically consistent with human body part structure.  
Therefore, current research \cite{chen2016semantic,ruan2019devil,xia2017joint,nie2018mutual} identifies human edge and pose as complementary cues to improve parsing performance. As shown in Fig.~\ref{fig:motivation}(b), when directly using generic segmentation methods for human parsing, some pixels of upper clothes are predicted as pants: the network incorrectly locates the edges between the two categories. Moreover, due to the lack of human semantic constraints, the left and right arms, left and right shoes are incorrectly identified. %, bringing about mistaken semantic body part topology. 
In Fig.~\ref{fig:motivation}(c), after adding edge information to parsing, we observe that the boundary pixels are accurately located. Further, when utilizing body part cues provided by pose features in Fig.~\ref{fig:motivation}(d), the mistaken prediction of the left arm no longer exists, and the left shoe is clearly distinguished from the right shoe. %From the second column, the network predicts upper body and lower body to the same category. %the network is able to segment left shoes beside right shoes.
	
In spite of the improvements so far, existing research utilising edge/pose to improve parsing has not leveraged them to the full potential. 
%Although edge or pose information could bring some gains to parsing task from different aspects, this line of research has two drawbacks. 
Usually a single factor, \emph{i.e.} either pose or edge, is used, which might be beneficial to handle \emph{a single problem} mentioned above. %usually studies a single factor, , and can only handle a single problem in parsing task, which is incomplete. 
In addition, existing methods typically perform feature concatenation or post processing for parsing refinement. We point out that this practice might be inferior. % to our strategy, \emph{i.e.,} using feature correlations. %thus is not effective to leverage sufficient contextual cues.
As shown in Fig.~\ref{fig:motivation}(c) and (d), when only a single factor is concerned for parsing system, there still remain blurs and holes in arm and dress area, and left/right shoes are inexactly predicted. Therefore, simple fusion or post processing may not be enough to process fine regions, such as edges of different parts.
To address this problem, we explore the correlations among edge and pose and find that it is preferable that edge, pose and parsing are simultaneously integrated.% three information.

%orthogonal
In this paper, we propose a Correlation Parsing Machine (CorrPM) to take advantage of both human semantic edge and pose features to benefit human parsing. Contrary to performing feature concatenation or post processing, we learn the correlation among the three tasks.  %contextual parsing machine that unifies parsing, edge and pose information into one model and exploit what advantages another two features could bring to parsing results.
	%On the basis of above observations, it is critical to explore an effective and efficient method correlating human semantic boundaries, pose and parsing to learn better parsing models. Therefore, we 
The CorrPM has three encoders and is featured by a heterogeneous non-local (HNL) module. The encoders calculate vector representations of the human edge, pose and semantics, respectively.  
	%Specifically, the proposed framework 
	%includes three encoders which encode the original image into parsing, pose and edge representations by three supervisions respectively. And then a heterogeneous non-local module is conducted to 
HNL mixes the three features into a hybrid representation and explores the spatial affinity between this hybrid feature and the parsing feature map at all positions.  
	%all positions between it and the parsing feature map. 
As such, our method can effectively perceive the human edges and maintains the integrity of a semantic region, addressing the inaccurate boundary localization problem. Meanwhile, by perceiving the body keypoints, our method improves the consistency of the body part geometry. %, especially the left and right limbs will be precisely distinguished.
For example, as shown in Fig.~\ref{fig:motivation} (e), our method corrects the mislabeling of arm region and correctly segments the boundary between upper clothes and pants, and between dresses and arms.% is correctly segmented. 
	
	%介绍contribution
To summarize, our contribution is three-fold.
1) We propose to use a Heterogeneous Non-Local (HNL) structure to capture the correlations among three closely related factors.
2) We show that human edge and pose, when both integrated in the Correlation Parsing Machine (CorrPM), bring significant improvement to parsing task.
3) Using simple edge detection and pose estimation models, we report very competitive parsing accuracy on three human parsing datasets.

	%------------------------------------------------------------------------

\section{Related Work}
	%Our work is related to the following areas, including semantic segmentation, human parsing and non-local network.
	\textbf{Semantic Segmentation.} Human parsing is a fine-grained semantic segmentation, which performs per-pixel prediction on all objects. Due to its great prospects in application, semantic segmentation has gained much importance in the past few years. FCN \cite{long2015fully,chen2016attention,zhao2017self} performs well on this task which applies fully convolution on the whole image to produce labels of every pixel. Inspired by this, many researchers \cite{Noh2015Learning,Badrinarayanan2015SegNet,Ronneberger2015U} start to leverage the encoder-decoder structure which extracts features by downsampling and then use upsampling to recover them to the original resolution. Aiming to enlarge the receptive field, another structure, DeepLab \cite{Chen2017deeplab}, designs atrous convolution kernels to force the network to perceive larger area and reduce the prediction errors. Zhao \textit{et al. }\cite{Zhao2016Pyramid} propose a pyramid scene parsing network aggregating multi-scale object clues to make the segmentation more precise. In \cite{xia2016zoom}, Xia \textit{et al.} propose the ``Auto-zoom Nets" to automatically ``zoom" the objects and parts which have diverse scales. 
	
	\textbf{Human Parsing.}
	Following main approaches in semantic segmentation, early researches in human parsing contribute towards this topic mostly by hand-crafted features and post-processed by Conditional Random Field (CRF) \cite{farabet2012learning,lin2016efficient}. Dong \textit{et al.} \cite{Dong2014A} use a variety of parselets assembled by ``And-Or" sub-trees to jointly parse human body labels and keypoint locations. 
	With the development of convolutional neural network (CNN), especially after the ResNet \cite{he2016deep} is proposed, many deep learning approaches have achieved much progress in this area. In \cite{liang2015human}, Liang \textit{et al.} propose a Co-CNN framework capturing cross-layer local and global context information to boost the parsing performance. Gong \textit{et al.} \cite{gong2017look} introduce a new large-scale benchmark LIP and a novel self-supervised structure-sensitive learning method. In \cite{li2017multiple}, Li \textit{et al.} tackle the human parsing problem by generating global parsing maps for person in a bottom-up way. 
	
	\textbf{Utilizing edge or pose for parsing.} Aiming to get more accurate predictions in human parsing task, recent works \cite{fang2018weakly,nie2018mutual,gong2018instance,gong2017look,xia2017joint,gong2019graphonomy,Dai2016Instance,Jian2014Towards,Nie2018Human} utilize edge or pose information as a guidance. 
	Chen \textit{et al.} \cite{chen2016semantic} propose an edge-aware filtering method to capture accurate semantic contours between two adjacent parts.
	%In addition to using keypoints, some previous works leverages edge detection as a prior to sharpen parsing details. 
	Ruan \textit{et al.} \cite{ruan2019devil} fuse the edge map with parsing feature which can reserve the boundary of person parts to benefit the human parsing. Gong \textit{et al.} \cite{gong2018instance} conduct both semantic part parsing and edge detection in the way of sharing intermediate representation of both features.  Xia \textit{et al.} \cite{xia2017joint} train two FCNs to predict poses and parts separately and then fuse them through a fully-connected conditional random field (FCRF) as a refinement.  Nie \textit{et al.} \cite{nie2018mutual} observe that pose and parsing can simultaneously boost the performance of each other by training two parallel models and adapt the mutual parameters.
	Despite the improvement, the existing methods simply perform feature concatenation or pose-processing to refine parsing results, which is inferior to guide parsing model to learn contextual cues. Our framework simultaneously integrates edge, pose and parsing representation and effectively exploits the correlation among these three representation. %Consequently, the overall network learns towards beneficial to human parsing. 
	
	\textbf{Non-local Network. }Human parsing is closely complementary to semantic edge information and human pose information. And the relationship among them is exploited  and employed by HNL which is modified from non-local network. Originating from non-local means algorithm \cite{Buades2005A} , the non-local network is leveraged in many approaches to capture long-range dependencies \cite{Zhu_2019_ICCV,zhang2019residual}. Wang \textit{et al.}\cite{wang2018non} propose the non-local block as a weighted summation of relationships of every position and show good performance in video classification. Even though non-local network has been a great success in many tasks , existing methods seek the relationship with the feature itself. Different from the existing self-attention mechanism, the proposed heterogeneous non-local module  aggregates parsing, edge and pose factors together and learns the correlation of parsing with the other two features.
	
	\begin{figure}[t]
	\begin{center}
		\centering
		%\fbox{\rule{0pt}{2in} }
		\rule{.9\linewidth}{0pt}
		\includegraphics[width=0.9\linewidth]{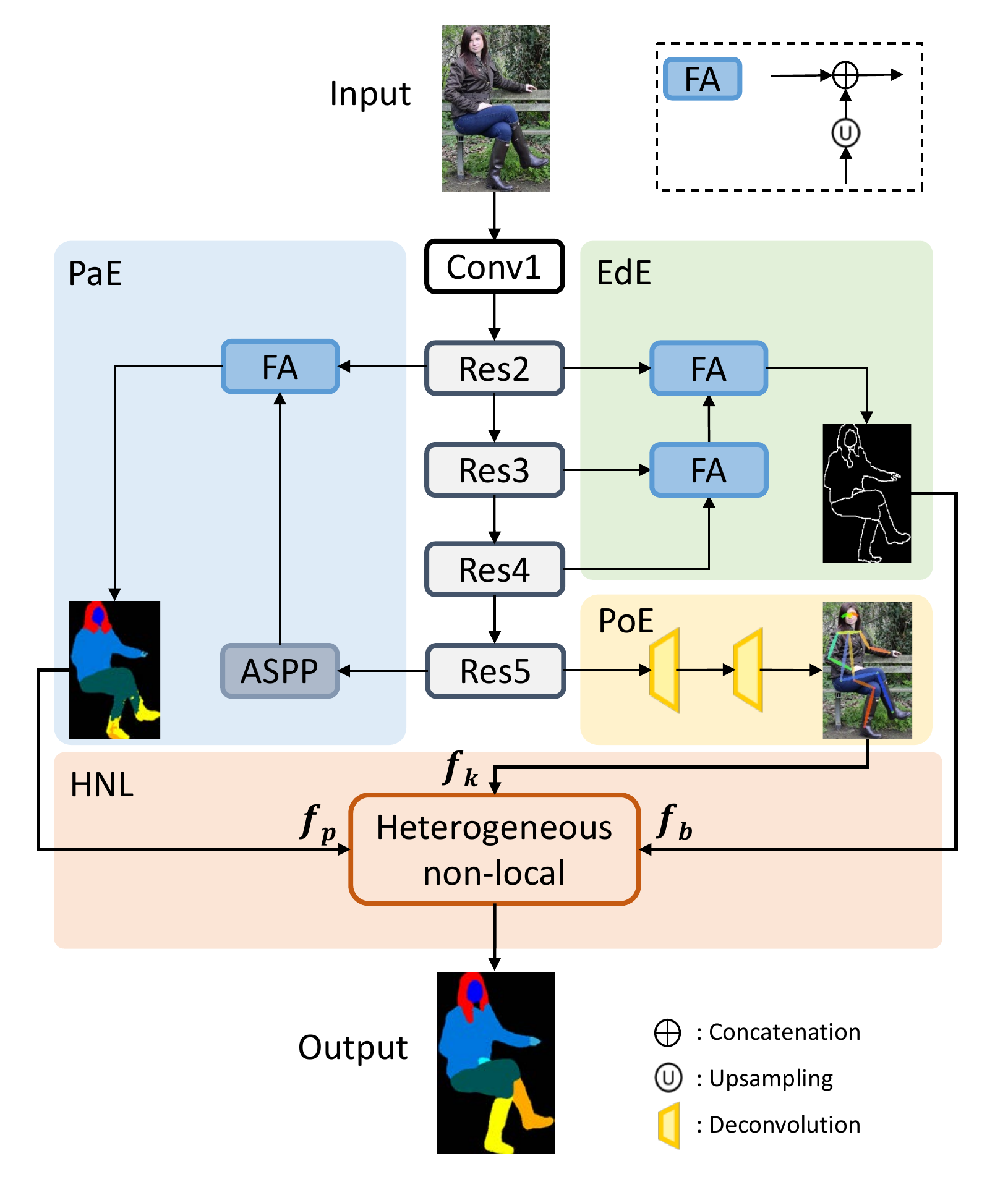}
	\end{center}
	\vspace{-3mm}
	\caption{Overview of the proposed network. PaE: parsing encoder. EdE: edge encoder. PoE: pose encoder. HNL: heterogeneous non-local module. FA: feature aggregation. $f_*$: parsing/edge/pose features. After extracted by three encoders, parsing, pose and edge features are fed into HNL to explore their correlation to benefit human parsing task.}
	\vspace{-3mm}
	\label{fig:framework}
	\end{figure}
	
\section{The Proposed Approach}
	
	As illustrated in Fig.~\ref{fig:framework}, the proposed Correlation Parsing Machine (CorrPM) leverages  human body keypoint and semantic boundary information to benefit human parsing. We firstly introduce the overall formulation of our framework in Section~\ref{subsection:formul}. The three feature encoders are represented in Section~\ref{subsection:enc} and we propose a heterogeneous non-local module (HNL) to correlate the three factors in Section~\ref{subsection:non_loc}. Then, Section~\ref{subsection:discuss}  explains the difference between the proposed HNL and the traditional non-local networks. And the overall training objective is illustrated in Section~\ref{subsection:train}. 
	%how these modules handle the two problems existing in human parsing as discussed earlier  
	
	\subsection{Formulation }\label{subsection:formul}
	Given an input image $I\in$  $\mathbb{R} ^ { 3 \times M \times N} $ of size $M \times N$, our task is to predict the label of every pixel and generate a segmentation mask $P \in \mathbb{R} ^ { M \times N } $ leveraging three kinds of information: human body part category $\mathcal{P}\in \{0,1,...,Q\}^{M\times N}$, semantic boundary $\mathcal{B}\in \{0,1\}^{M\times N}$ and human body keypoint location $\mathcal{K}=\{(x_i, y_i)\}_{i=1}^{J}$. $J$ and $Q$ are the number of body joints and part categories. $(x_i, y_i)$ are the coordinates of the point $i$, and the pixels that belong to boundaries are labeled as 1 otherwise 0. We aim to design a unified framework that jointly utilizes these three factors and uncovers the correlation among them to better leverage the pivotal contextual cues. 
	%Our model is formulated as following:
	%\begin{equation}
	%S = Corr(E_{seg},E_{edge},E_{pose}) , 
	%\label{eq:rela1}
	%\end{equation}
	%where $E_*$ indicates parsing, edge and pose features extracted by their encoders respectively. $Corr(*)$ is the heterogeneous  correlation block discovering the spatial affinity among feature maps from theboundary, pose and parsing.
	
	\subsection{Feature Encoding}\label{subsection:enc}
	Human parsing, pose estimation, edge detection are complementary and closely related, hence, their features can be learned by a shared base model $\Theta$, \eg, ResNet101 \cite{he2016deep}. The feature at the lower stage of the base model retains high resolution structure and is fed to edge encoder to capture the object edge boundaries $f_{b}$. And the higher-stage feature keeps rich semantic information which is further used as parsing feature $f_ {p}$ and keypoint feature $f_ {k}$. 
	
	\textbf{Parsing Encoder.} 
	We adopt a parsing pipeline to predict a coarse segmentation map firstly. Context information is leveraged in many previous work \cite{Zhao2016Pyramid,Chen2017deeplab} in semantic segmentation and it is also essential in human parsing. Given the parsing feature of the base model $\Theta$, we observe that merely performing dense pixel-wise prediction on it will cause mislabeling. Therefore, we add the Atrous Spatial Pyramid Pooling (ASPP) \cite{Chen2017deeplab} to enlarge the receptive fields and get more useful context cues. 
	%It consists of five parallel branches: a global average pooling layer, a $1 \times 1$ convolution layer and three $3 \times 3$ atrous convolution layers with rates of (12, 24, 36). After that, all feature maps are concatenated and the channels are reduced by $1 \times 1$ convolution. 
	%Furthermore, some distinct parts while have similar appearances are hard to distinguish if global contextual information is not provided. 
	%The global average pooling layer is utilized to capture image-level content, which is followed by a $1 \times $1 convolution, then the feature is upsampled to the same size as the original feature map extracted by ResNet101. The four convolution branches can get multi-level context of one pixel and dilation rate is (1, 12, 24, 36), respectively. 
	
	Meanwhile, some objects in human parsing have quite low resolution, \eg, sunglasses and socks, so the details might be lost in the process of downsampling. We employ the feature of \emph{Res2} of base model and upsample the output of ASPP module to the same scale as \emph{Res2} and concatenate them as $f_{p}$.
	%\textbf{high resolution} Different from general semantic segmentation, some parts in human parsing could be quite small such as shoes, socks and sunglasses. Especially after going through several convolution layers, although the feature has a strong semantic consistency, its resolution will get too low  to recover the original information. 
	%It is observed that the feature map in low stage of network retains large spatial scale so we guide the prediction in high stage by combining with low stage features. 
	%As mentioned above, ResNet is adopted as the base network which has five stages, \textit{e.g.} res1-5. And we use the feature of res2 as the guidance cue. The output of ASPP module is upsampled by bilinear interpolation and then concatenated with the guidance feature after stage2. In this way, it keeps both high-level semantic context and high-resolution structure which would be input to \textbf{later correlation operation}. %At last, this fusion feature is fed into a classification layer supervised by .
	After extracted from the parsing encoder, the feature $f_{p}$ obtains a coarse semantic representation and will be further fed into the heterogeneous non-local module to obtain pose and edge guidance. % for a dense parsing prediction.
	
	\textbf{Pose Encoder.}
	In order to get human body structure cues, we design a pose encoder to get joint locations. Many existing approaches \cite{xia2017joint,nie2018mutual,newell2016stacked} in pose estimation adopt complicated CNNs to get more accurate keypoint locations. For instance, Hourglass \cite{newell2016stacked} performs repeated downsampling and upsampling procedure to capture multi-scale keypoint information. Different from them, we only deploy two transposed convolution layers \cite{xiao2018simple} to extract human keypoint structure, since pose estimation task can also get benefits from the parsing task. %And it needs no pre-training. The kernel sizes of this two layers are both set as 4 with stride 2. 
	As a result, the shared feature is upsampled by 4 times generating the pose feature $f_{k}$. It is the same scale as the parsing feature $f_{p}$.
	
	After the pose representation $f_{k}$ is captured, we regress the heatmap from it. Following \cite{sun2019deep}, we apply 2D Gaussian filter centered on each annotated keypoint coordinate with standard deviation of 7 pixels, and generate the ground truth heatmap as the supervision of pose encoder. 
	%Mean Square Error is used as the pose loss function $L_{pose}$ to close the gap between the predicted heatmap and the ground truth gaussian heatmap.
	
	\textbf{Edge Encoder.}
	In human parsing task, semantic boundary ambiguity remains to be solved. The border pixels of two adjacent semantic parts may be inaccurately predicted, particularly when they have similar appearances. %Thus, there is a demand to strengthen the boundary between confusing parts. 
	Hence, we propose an edge encoder to learn feature $f_{b}$ with boundary consciousness. It is observed that lower stages in neural network maintain high resolution and higher-stage feature obtains detailed semantic information. As shown in Fig.~\ref{fig:framework}, we leverage the features of  \emph{Res2},  \emph{Res3} and  \emph{Res4} which retain both large spatial details and semantic consistency. The feature maps are upsampled to the same size as \emph{Res2} by linear interpolation. Then, they are concatenated and fed into a $1 \times 1$ convolution layer to generate the edge feature map $f_{b}$. The edge encoder is supervised by the edge information between two adjacent categories and the feature will be further fed into the heterogeneous non-local correlation block. %extracted from the ground truth mask. 
	
	\begin{figure}[t]
	\begin{center}
		\centering
		%\fbox{\rule{0pt}{2in} }
		\rule{.9\linewidth}{0pt}
		\includegraphics[width=1\linewidth]{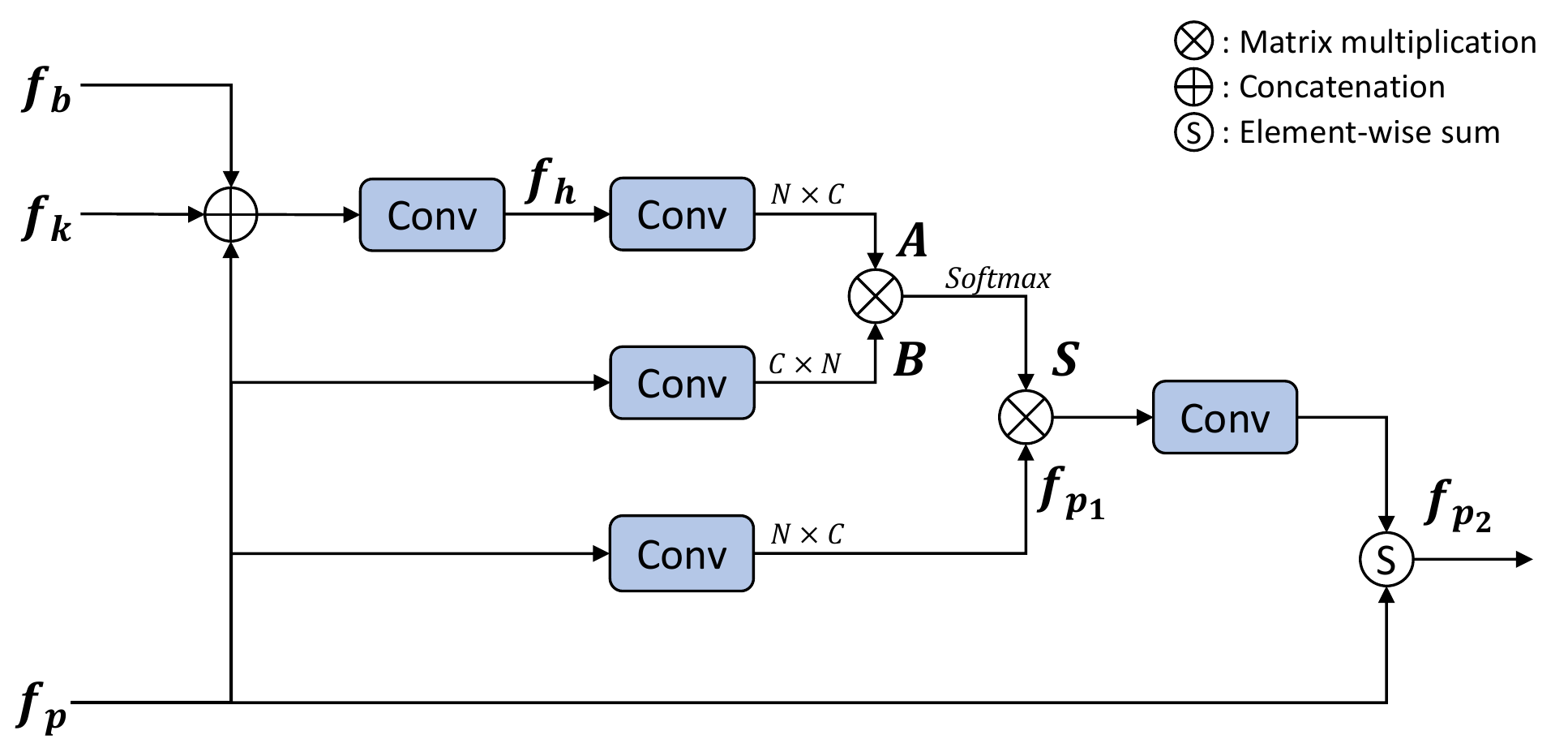}
	\end{center}
	\vspace{-4mm}
	\caption{Structure of the heterogeneous non-local (HNL) module. It aggregates parsing, edge and pose feature into a hybrid feature $f_h$ and calculates the correlation between $f_h$ and $f_p$.} 
	\vspace{-3mm}
	\label{fig:HNL}
	\end{figure}	

	\subsection{Heterogeneous Non-Local }\label{subsection:non_loc}
	
	Many existing researches prove that either edge or pose is a beneficial factor to parsing task. However, the fusion strategy they employ cannot fully leverage the two factors as discussed above. Recently, the correlation module is used to capture the long-range contextual information by self-attention \cite{fu2018dual,Huang_2019_ICCV} or explore the relationship between the two features \cite{Zhu_2019_ICCV}. 
	%If we follow this routine practice, an intuitionistic solution to our parsing task is to calculate the relation matrix between each two feature maps, then apply each relation matrix to the parsing feature for more dense segmentation result. 
	However, if we follow this operation, the correlation computation cost is high ($O(n^2)$, $n$ is the number of feature maps) and the overall model is hard to converge.
	Therefore, we propose a Heterogeneous Non-Local (HNL) block to fully leverage the contextual cues provided by boundaries and poses, which we believe is more effective and more efficient. 
	
	%1.卷积
	%\textbf{Aggregation module } 
	As shown in Fig.~\ref{fig:HNL}, we first aggregate the three factors by concatenating them in the channel dimension, and then a convolution layer parameterized by $W_{a}$ is conducted to transform it into a hybrid feature $f_ {h}$, whose dimension is the same as that of $f_{p}\in\mathbb{R}^{C\times H\times W}$ :
	\begin{equation}
	f_h =  W_{a}(f_{p}\oplus f_{b}\oplus f_{k}), 
	\label{eq:conv}
	\end{equation}
	where $\oplus$ means concatenation.

	We exchange the self-attention in the standard non-local block with correlation between the hybrid feature $f_{h}$ and parsing feature $f_{p}$. First, $f_{h}$ and $f_{p}$ are fed into two convolution layers to generate two new features $A$ and $B$, then we reshape them into matrixes with size $N \times C$ and $ C \times N $, respectively, where $N = H \times W$ denotes the total number of pixels per channel. We compute the relationship map $\textbf{\emph{S}}$ ${\in \mathbb{R} ^ {N \times N} }$   by a matrix product of $A$ and $B$, and normalize the relation map by a softmax operation.  
	\begin{equation}
	%s_{i,j} = \frac{exp(X_i \cdot Y_j)}{\sum_{j=1}^N exp(X_i \cdot Y_j)},
	\textbf{\emph{S}} = softmax(A \cdot B)
	\label{eq:rela1}
	\end{equation}
	where  a point ${(i,j)}$ in $\textbf{\emph{S}}$ measures the relation affinity between the $i^{th}$ pixel in hybrid feature $f_{h}$ and $j^{th}$ pixel in parsing feature $f_{p}$. %$A$ and $j^{th}$ pixel in $B$. %, and $Z(i) = \sum_{\forall j}S(i,j)$ is the sum of the similaries of $i$ and all pixels belonging to parsing feature. 
	
	Then we feed the parsing feature $f_{p}$ into another convolution layer to generate $f_{p_1}\in \mathbb{R} ^ {C\times H \times W}$ and reshape it to $\mathbb{R} ^ {N \times C}$, which is multiplied by $\textbf{\emph{S}}$ to integrate the pixel correlation cues into parsing features. The resulting feature is fed into the final convolution layer parameterized by ${W_b}$ and added back to $f_{p}$ element-wise to get the final parsing feature $f_{p_2}$. The overall procedure can be formulated as:
	\begin{equation}
	%\overline f^{p}_{i} =  {W_b}\sum_{j=1}^{N}(s_{i,j}Z_j) + f_{i}^{p} ,
	f_{p_2} = W_b (\textbf{\emph{S}}\cdot f_{p_1}) + f_p,
	\label{eq:rela2}
	\end{equation}	
	where ${W_b}$ is initialized as 0. In this way, the hybrid representation effectively aggregates parsing, edge and pose information together. And the refined parsing feature $f_{p_2}$ in Equation \ref{eq:rela2} is a weighted summation of every position in the hybrid feature and the parsing feature. Therefore, it obtains edge information between two bordered parts and retains semantic consistency with human body, thus getting more reasonable parsing results.
	
	%\begin{figure}
	%	\begin{center}
	%		\rule{.9\linewidth}{0pt}
	%		\includegraphics[width=1\linewidth]{Figures/heatmap}
	%	\end{center}
	%	\vspace{-4mm}
	%	\caption{Visualization of the relation map of a point(marked by red cross) on an image from LIP validation set. The  third and fourth column are two parsing predictions w/o HNL module.  }
	%	\label{fig:demo}
	%\end{figure}

	\subsection{Discussions}\label{subsection:discuss}
	%\textbf{The superiority of the correlation block to non-local networks. }
	The heterogeneous non-local block is an extension of non-local neural network \cite{wang2018non}. However, different from the traditional non-local operation which only computes the relationship of one feature as a mechanism of self-attention, the proposed network has three advantages. First, it integrates human parsing, pose estimation and edge detection tasks into a unified model, and the correlation is calculated among three different feature representations. Second, HNL does not add much computation complexity compared with traditional non-local structure while maintains very competitive accuracy. %instead of calculating the relationships between any two of the three, which is not effective and efficient, the proposed framework  aggregates these three features into a hybrid representation. Accordingly, the heterogeneous non-local block fully exploits these three sets of information and the overall framework could learn those that are beneficial to human parsing. 
	Finally, for other tasks which are also related to human parsing, it has potential to integrate it into the hybrid representation and model the relationship among them with only little computation complexity increased (brought by the corresponding encoder). %without increasing computation complexity.
	
	%这里的finally本来想问下您，这样写合不合适的，忘了 。。。
	%\begin{figure}
	%	\begin{center}
	%		\rule{.9\linewidth}{0pt}
	%		\includegraphics[width=1\linewidth]{Figures/nl}
	%	\end{center}
	%	\vspace{-2mm}
	%	\caption{Illustration of non-local  correlation block. ``$\otimes$'' denotes matrix multiplication, ``$\oplus$'' denotes element-wise addition, and "R" is reshape operation, "C" denotes convolution layer. OP??.}
	%	\label{fig:Correlation}
	%\end{figure}
	
	\subsection{Training objectives} \label{subsection:train}
	In addition to the parsing supervision, human keypoint location and semantic edge information are utilized to train the whole model. %As mentioned above, after encoded by respective encoders, $f_{p}$, $f_{k} $ and $f_{b} $ are fed into classification layers $\hat{C}_m$, $m\in$ \{1, 2, 3, 4\} to predict different results. 
	The total training objective is: % defined as follows:
	\begin{equation}
	%\begin{aligned}
	%L &= \alpha (L^{p}(\hat{C}_1(f^{''}_{p}), \mathcal{P}) + L^{p}(\hat{C}_2(f_{p}) , \mathcal{P})) \\
	%&+\beta L^{b}(\hat{C}_3(f_{b}),\mathcal{B}) + \gamma L^{k}(\hat{C}_4(f_{k}), \mathcal{K})
	L = L_{p_2}+L_{p} + \alpha L_{b} + \beta L_{k},
	\label{eq:loss}
	%\end{aligned}
	\end{equation}
	%where $\mathcal{P}$, $\mathcal{B}$ and $\mathcal{K}$ are parsing, edge and pose annotations. 
	$L_{p_2}$ or $L_{p}$ is the loss between the parsing result $f_{p_2}$ or $f_p$ and the parsing annotations; %$L_{p}$ denotes the loss between the encoded feature $f_p$ and the parsing label;
	$L_b$ denotes the loss between the predicted edge map $f_b$ and the edge annotation; $L_{k}$ is the loss between the body joints prediction $f_k$ and the ground truth coordinates.
	It is worth noting that the edge annotation is obtained by finding the borders of the mask between two different semantic parts, which needs no additional annotations. 
	Cross Entropy loss is adopted as $L_{p_2}$, $L_{p}$ and $L_{b}$, and Mean Square Error loss is used for $L_{k}$. The whole framework is trained end-to-end.

	%------------------------------------------------------------------------
	\section{Experiments} \label{section:exper}
	\subsection{Experimental Settings}
	\textbf{Datasets and metrics. }We evaluate the performance of the proposed method  on three human parsing datasets: 
	
	LIP \cite{gong2017look} is a large-scale benchmark dataset focusing on semantic understanding of human body parts and clothes labels. It contains %both human pose and parsing groundtruth, including 
	coordinates of 16 body keypoints and pixel-level annotations of 20 semantic human parts (including one background label). There are totally 50,462 images which are further split into train/val/test sets containing 30,462/10,000/10,000 images, respectively.
	
	ATR \cite{liang2015human} contains 18 categories of human part labels including \emph{face, sunglasses, hat, scarf, hair, upper-clothes, left/right arm, belt, pants, left/right leg,  skirt, left/right shoe, bag, dress} and \emph{background}.
	Following \cite{liang2015human}, we use 16,000 images for training, 1,000 for testing and 700 for validation. 
	
	CIHP \cite{gong2018instance} provides 38,280 images with 20 categories. It contains 28,280 training, 5,000 validation and 5,000 test images.
	On account of no human pose annotations in ATR and CIHP, we utilize the pose estimator \cite{xiao2018simple} trained on COCO  \cite{lin2014microsoft} to obtain human body keypoint locations as ground truth. 
	%It contains 17 body joints. including \emph{nose, left-eye, right-eye, left-ear, right-ear, left-shoulder, right-shoulder, left-elbow, right-elbow, left-wrist, right-wrist, left-hip, right-hip, left-knee, right-knee, left-ankle, right-ankle}.
	%MHP v2.0\cite{li2017multiple} captures real-world multi-person in crowded scenes. It provides 16 body joint categories and 59 semantic part segmentation labels of human, including 15,403 images for training, 5,000 for validation and 5,000 for testing. 
	During training, we first utilize Mask R-CNN\cite{he2017mask} to generate the mask of every person, and apply it on multi-person images to generate single person images. We obtain 93,213 training images in total. During inference, single person is segmented from background in the same way as training and we conduct parsing with the proposed network and finally merge them into the original image.

	%PASCAL-Person-Part is a subset of PASCAL VOC 2010 providing additional pose joint locations and semantic part segmentation labels of human. There are 14 body joints and 6 body parts labels including head, torso, upper/lower arms, and upper/lower legs. And it includes 1,716 images for training and 1,817 for testing.
	
	%We report Pixel Accuracy, Mean Accuracy and mIoU to evaluate the human parsing performance on LIP. For CIHP, the performance is validated in terms of mIoU. For ATR dataset, we use the same metric as \cite{liang2015human}, including Pixel Accuracy, Foreground Accuracy, Average Precision, Average Recall and F-1 score.
	We report Accuracy, mIoU, Precision, Recall and F-1 score to evaluate the parsing performance on the datasets.

	\begin{table}[]
		\centering
		\footnotesize %\scriptsize
		\begin{tabular}{ccccc}
			\hline
			%Method & Pixel Acc. & Mean Acc. & mIoU \\ \hline
			Method & EA & BI & Fusion Strategy & Accuracy \\ \hline 
			\cite{ruan2019devil} & \checkmark & & Feature concatenation & ++ \\
			\cite{gong2018instance} & \checkmark & & Feature concatenation & ++ \\
			\cite{nie2018mutual} & & \checkmark & Parameters mutual learning & ++ \\ 
			\cite{gong2017look} & & \checkmark  & Loss constraint & + \\
			\cite{xia2017joint} & & \checkmark  & Post processing & + \\ \hline
			Ours &  \checkmark  & \checkmark  & Correlation & +++ \\ 
			\hline
		\end{tabular}
		\vspace{1mm}
		\caption{Comparison of different fusion methods. EA represents edge ambiguity and BI represents boundary inconsistency. Existing methods use either edge or pose to solve a single problem in parsing. Different from them, we aggregate parsing, edge and pose feature and explore the correlation among them which shows the superiority on Accuracy.}
		\vspace{-3mm}
		\label{tab:summary}
	\end{table}

	\textbf{Training Details. }We train CorrPM from scratch for 150 epochs, and adopt ResNet101 \cite{he2016deep} pre-trained on ImageNet \cite{deng2009imagenet} as the base model $\Theta$. %The stride of $\Theta$ is 16, \textit{i.e.}, the scale of output feature map is 1/16 of that of the input image. 
	During training, the $384 \times 384$ input images are randomly rotated (from {${-60}^{\circ}$} to {${60}^{\circ}$}), flipped and resized (from 0.75 to 1.25). $f_{p}$, $f_{k} $ and $f_{b}$ are in the same size of $C\times H \times W$, where $C=512$ and $H=W=96$. 
	%Every of the classifiers $\hat{C}_m$, $m\in$ \{1, 2, 3, 4\} is implemented by $1 \times 1$ convolution layer.
	We use SGD as the optimizer and the learning rate is initially set as 1e-3. Following previous works \cite{DRN}, we employ the ``poly" learning rate policy, and the learning rate is multiplied by $(1-\frac{iter}{total\_{iter}})^{0.9}$. We set the momentum to 0.9 and weight decay to 5e-4. The edge loss weight $\alpha$ and pose loss weight $\beta$ are 2 and 70. %All edge annotations are obtained from the parsing labels by calculating the border from two adjacent body part categroies.
	
	\textbf{Testing phase.}
	During inference, the outputs of pose and edge branches are ignored, and $f_{p_2}$ is employed to predict the final  parsing mask $P$. The inference procedure is executed on a 12GB TITAN V  for a fair speed comparison with other methods. Our model does not add too much complexity compared with direct concatenation, because the base model (ResNet-101) consumes a majority of computations. CorrPM achieves a speed of 11 fps which is faster than Attention+SSL \cite{gong2017look} (2 fps) and MuLA \cite{nie2018mutual} (5 fps).
	%For data augmentation, all training images go through a series of  operations including rotation (from {${-60}^{\circ}$} to {${60}^{\circ}$}), flipping and scaling (from 0.75 to 1.25).

	%\renewcommand{\arraystretch}{1.2}
	
	\subsection{Comparison with related methods}
	\subsubsection{Fusion strategy comparison}
	%\vspace{-2mm}
	Tab.~\ref{tab:summary} lists some existing researches that utilize pose or edge information to assist human parsing task. For edge ambiguity issue, \cite{gong2018instance} and \cite{ruan2019devil} extract edge feature and concatenate it with parsing feature to perceive useful cues of part boundaries. But this fusion strategy is not able to sufficiently obtain the semantic boundary completeness.
	Aiming to solve body inconsistency problem, \cite{nie2018mutual} conducts two parallel human pose estimation and human parsing networks and mutually learns the parameters. However, the training process is somewhat complicated. Meanwhile, \cite{xia2017joint} adopts FCRF as a way of post-processing and \cite{gong2017look} adds a joint loss utilizing the pose information to constrain part segments. Above fusion methods only employ a single factor and merely handle a single problem. In comparison, our CorrPM combines parsing with pose and edge information, and the experiment also shows exploring the correlation among the three factors is a superior feature fusion strategy to other recent methods.
	
	\subsubsection{Performance on single-person datasets}
	
	\begin{table}[t]
		\centering
		\footnotesize
		\begin{tabular}{cccc}
			\hline
			Method & Pixel Acc. & Mean Acc. & mIoU \\ \hline
			DeepLabV2 \cite{Chen2017deeplab} & 82.66 & 51.64 & 41.64 \\
			Attention \cite{chen2016attention} & 83.43 & 54.39 & 42.92 \\
			Attention+SSL \cite{gong2017look} & 84.36 & 54.94 & 44.73 \\
			SS-NAN \cite{zhao2017self} & 87.59 & 56.03 & 47.92 \\
			MuLA(Hourglass) \cite{nie2018mutual} & \textbf{88.50} & 60.50 & 49.30 \\
			JPPNet \cite{Liang2018Look} &  86.39 & 62.32 &  51.37 \\  
			CE2P \cite{ruan2019devil} & 87.37 & 63.20 & 53.10 \\ \hline
			Ours$\dagger$ & 87.36 & 66.37 & 54.43 \\
			Ours & 87.68 & \textbf{67.21} & \textbf{55.33}\\
			\hline
		\end{tabular}
		\vspace{1mm}
		%没看清，意思是去掉冠词吗
		\caption{Comparison of different methods on the validation set of the LIP dataset. $\dagger$ means removing $L_p$ in Equation ~\ref{eq:loss}.}
		\label{tab:iouLIP}
		\vspace{-1mm}
	\end{table}
	
	\begin{table}[]
	\centering
	\footnotesize %\scriptsize
	\begin{tabular}{cccccc}
		\hline
		Method & Acc & F.g.Acc & Pre & Rec & F-1 score \\ \hline
		DeepLabV2 \cite{Chen2017deeplab} & 94.42 & 82.93 & 69.24 & 78.48 & 73.53 \\
		Attention \cite{chen2016attention} & 95.41 & 85.71 & 81.30 &  73.55 & 77.23 \\
		CoCNN \cite{liang2015human} & 96.02 & 83.57 & 84.59 & 77.66 & 80.14  \\
		TGPNet \cite{luo2018trusted} & 96.45 & 87.91 & 83.36 & 80.22 & 81.76 \\  \hline
		Ours & \textbf{97.12} & \textbf{90.40} & \textbf{89.18} & \textbf{83.93} & \textbf{86.12} \\ 
		\hline
	\end{tabular}
	\vspace{1mm}
	\caption{Comparison of Accuracy, Foreground Accuracy, Precision, Recall and F-1 score on the ATR test set.}
	\vspace{-3mm}
	\label{tab:iouATR}
	\end{table}	
	
	\textbf{LIP. }
	We show the performance comparison of the proposed model and the other methods on LIP validation set. As shown in Tab.~\ref{tab:iouLIP}, the proposed CorrPM achieves the best performance of 55.33\% in terms of mIoU  and significantly outperforms other methods. Specifically, JPPNet and MuLA add pose supervision as a constraint of human parsing. CE2P adds edge information to refine parsing results. Their experiment results show that pose and edge cues help achieve better performance. However, the pose or edge in formations are not fully exploited. By exploring the correlation between the three factors, the HNL brings a boost of 2.23\% mIoU to CE2P and 6.03\% mIoU to MuLA. Even when removing the loss $L_p$, the 54.43\% mIoU is higher than others, which indicates the direct supervision of the parsing encoder is necessary and our framework effectively utilizes pose and edge features to assist human body parsing.
	Moreover, the pose encoder in our network only consists of two deconvolution layers, and it is much simpler than the hourglass which is adopted in MuLA \cite{nie2018mutual}. Thus, the performance may get higher if utilizing more powerful network.

	\textbf{ATR. }
	Tab.~\ref{tab:iouATR} reports the results and comparisons with four recent approaches on ATR. The proposed method brings a significant performance gain in terms of every metric. Particularly, our model achieves 4.36\% boost for F-1 score. This increase confirms the effectiveness of the pose and edge factors to parsing, and the correlation module has a strong capability to incorporate pose and edge information with the parsing features. %Note that we do not apply multi-scale feature pyramid fusion \cite{luo2018trusted}. 
	Although the F-1 score 90.89\% in \cite{gong2019graphonomy} is higher than ours, it adopts DeepLabV3+ as backbone which is more complicated than ResNet101, and the input size $512\times 512$ is larger than our $384\times 384$. 
	On the basis that the human joint labels are obtained from the output of the pose estimator \cite{xiao2018simple}, it illustrates that the proposed system is flexible and has a low-complexity to be deployed with no additional pose annotation cost.

	\begin{table}[]
		\centering
		\small %small
		\begin{tabular}{cccc}
			\hline
			%Method & Pixel Acc. & Mean Acc. & mIoU \\ \hline
			Method & Backbone & mIoU \\ \hline
			PGN \cite{gong2018instance} & ResNet101 & 55.80 \\
			Parsing R-CNN (R50) \cite{yang2019parsing} & ResNet50 & 57.50 \\
			Graphonomy \cite{gong2019graphonomy} & DeepLabV3+ & 58.58 \\
			Parsing R-CNN (X101) \cite{yang2019parsing} & ResNeXt101 & 59.80 \\
			\hline
			Ours & ResNet101 & \textbf{60.18} \\ 
			\hline
		\end{tabular}
		\vspace{1mm}
		\caption{Comparison of performance on the CIHP validation set.}
		\label{tab:iouCIHP}
		\vspace{-3mm}
	\end{table}

	\begin{table*}[]
		\centering
		\scriptsize
		%\resizebox{\textwidth}{10mm}{
		\setlength{\tabcolsep}{1mm}{
			\begin{tabular}{llllllllllllllllllllll}
				%{c{0.1cm}c{0.1cm}c{0.1cm}c{0.1cm}c{0.1cm}c{0.1cm}c{0.1cm}c{0.1cm}c{0.1cm}c{0.1cm}c{0.1cm}c{0.1cm}c{0.1cm}c{0.1cm}c{0.1cm}c{0.1cm}c{0.1cm}c{0.1cm}c{0.1cm}c{0.1cm}c{0.1cm}c{0.1cm}}	
				%{llllllllllllllllllllll}
				\hline
				Method & hat & hair & glove & glass & u-clot & dress & coat & sock & pants & j-suit & scarf & skirt & face & l-arm & r-arm & l-leg & r-leg & l-shoe & r-shoe & bkg & Avg \\ \hline
				%检查一下 又改了一点点  %好的看到了 交了算了 
				%没时间了 别超页 ok 祝好运 88
				%我提交了，现在在check
				%页数正好，我已经很知足了，感谢老师！！！您辛苦！！！！
				Attention \cite{chen2016attention} & 58.87 & 66.78 & 23.32 & 19.48 & 63.20 & 29.63 & 49.70 & 35.23 & 66.04 & 24.73 & 12.84 & 20.41 & 70.58 & 50.17 & 54.03 & 38.35 & 37.70 & 26.20 & 27.09 & 84.00 & 42.92\\
				DeepLabV2 \cite{Chen2017deeplab} & 56.48 & 65.33 & 29.98 & 19.67 & 62.44 & 30.33 & 51.03 & 40.51 & 69.00 & 22.38 & 11.29 & 20.56 & 70.11 & 49.25 & 52.88 & 42.37 & 35.78 & 33.81 & 32.89 & 84.53 & 44.03 \\
				MMAN \cite{Luo2018Macro} & 57.66 & 65.63 & 30.07 & 20.02 & 64.15 & 28.39 & 51.98 & 41.46 & 71.03 & 23.61 & 9.65 & 23.20 & 69.54 & 55.30 & 58.13 & 51.90 & 52.17 & 38.58 & 39.05 & 84.75 & 46.81 \\
				SS-NAN \cite{zhao2017self} & 63.86 & 70.12 & 30.63 & 23.92 & \textbf{70.27} & 33.51 & 56.75 & 40.18 & 72.19 & 27.68 & 16.98 & 26.41 & 75.33 & 55.24 & 58.93 & 44.01 & 41.87 & 29.15 & 32.64 & 88.67 & 47.92 \\
				JPPNet \cite{Liang2018Look} & 63.55 & 70.20 & 36.16 & 23.48 & 68.15 & 31.42 & 55.65 & 44.56 & 72.19 & 28.39 & 18.76 & 25.14 & 73.36 & 61.97 & 63.88 & 58.21 & 57.99 & 44.02 & 44.09 & 86.26 & 51.37 \\  
				CE2P \cite{ruan2019devil} & 65.29 & \textbf{72.54} & 39.09 & \textbf{32.73} & 69.46 & 32.52 & 56.28 & \textbf{49.67} & 74.11 & 27.23 & 14.19 & 22.51 & \textbf{75.50} & 65.14 & 66.59 & 60.10 & 58.59 & 46.63 & 46.12 & 87.67 & 53.10 \\ \hline
				P & 63.61 & 69.18 & 36.25 & 27.68 & 67.23 & 31.80 & 53.69 & 43.45 & 71.75 & 28.76 & 14.33 & 24.39 & 72.33 & 57.76 & 60.74 & 47.80 & 47.38 & 34.18 & 34.90 & 86.22 & 48.67 \\
				PP & 62.60 & 68.47 & 35.78 & 27.36 & 65.16 & 27.78 & 51.50 & 41.60 & 70.42 & 29.60 & 17.11 & 21.50 & 71.69 & 59.46 & 62.11 & 50.80 & 50.75 & 37.76 & 40.03 & 85.69 & 48.86 \\ 
				P+B & 65.11 & 70.71 & 38.38 & 30.04 & 68.65 & 32.60 & 55.13 & 46.31 & 73.37 & 31.94 & 17.51 & 28.36 & 73.51 & 60.68 & 63.52 & 51.50 & 51.37 & 39.75 & 39.78 & 87.09 & 51.27 \\
				P+K & 64.30 & 70.24 & 39.10 & 28.85 &  68.03 & 33.10 & 55.16 & 46.74 & 72.99 & 27.57 & 16.59 & 28.44 & 73.03 & 60.60 & 63.34 & 51.22 & 51.42 & 38.68 & 39.40 & 86.90 & 50.79 \\
				P+B+K & 65.01 & 71.13 & 40.30 & 29.14 & 69.47 & 33.91 & 55.78 & 47.82 & 73.85 & 31.98 & 18.81 & 28.94 & 74.12 & 61.93 & 63.95 & 52.35 & 51.99 & 40.19 & 40.81 & 87.23 & 51.93\\
				PB & 65.43 & 71.77 & 40.69 & 26.00 & 69.32 & 32.82 & 56.33 & 46.61 & 74.52 & 30.87 & 23.46 & 27.51 & 74.28 & 64.23 &  66.68 & 57.64 & 56.72 & 44.80 & 44.80 & 87.77  & 53.11 \\
				PK & 66.16 & 72.06 & 40.52 & 31.15 & 69.74 & 33.97 &  56.81 & 49.22 & 74.74 & \textbf{32.56} & 20.19 & 27.81 & 74.78 & 65.48 & 67.45 & 59.48 & 58.41 & 45.41 & 45.95 & 87.72 & 53.98 \\
				PBB & 66.14 & 72.42 & 41.04 & 27.81 & 70.12 & 34.91 & 57.01 & 47.21 & 75.03 & 31.38 & 22.99 & 28.21 & 74.39 & 64.92 & 67.58 & 58.33 & 57.64 & 45.51 & 46.10 & 87.46 & 53.82 \\
				PKK & 66.15 & 72.26 & 40.78 & 31.34 & 69.94 & 34.02 &  57.40 & 49.41 & 74.91 & 32.19 & 21.77 & 28.11 & 74.98 & 65.38 & 67.55 & 59.66 & 58.62 & 45.58 & 46.01 & 87.32 & 54.17 \\
				Ours (CorrPM) & \textbf{66.20} & 71.56 & \textbf{41.06} & 31.09 & 70.20 & \textbf{37.74} & \textbf{57.95} & 48.40 & \textbf{75.19} & 32.37 & \textbf{23.79} & \textbf{29.23} & 74.36 & \textbf{66.53} & \textbf{68.61} &  \textbf{62.80} & \textbf{62.81} & \textbf{49.03} & \textbf{49.82} & \textbf{87.77} & \textbf{55.33} \\ 
				\hline
			\end{tabular}
		}
		%\hspace{2mm}
		\vspace{1mm}
		\caption{Comparison of per-class IoU on the LIP validation set. P: Only parsing feature; PP: Performing self-correlation on parsing feature; P+B/P+K: Concatenating parsing with edge/pose feature; P+B+K: Concatenating parsing, edge and pose feature; PB/PK: Correlating parsing with edge/pose feature; PBB/PKK: Correlating parsing with two edge/pose features. CorrPM outperforms existing methods and achieves 55.33\% mIoU.}
		\vspace{-3mm}
		\label{tab:perIoU}
	\end{table*}

	\subsubsection{Performance on multi-person datasets}
	\textbf{CIHP. }
	Experiment results are compared with the other approaches in Tab.~\ref{tab:iouCIHP} on the CIHP dataset. Our model outperforms the existing approaches and achieves 60.18 in terms of mIoU. The previous work \cite{gong2018instance} gets 55.80\% mIoU by jointly conducting human parsing and edge detection. Parsing R-CNN \cite{yang2019parsing} gets 57.50\% mIoU using ResNet50 and its training images are in the size of $512\times 864$. Using smaller input size and backbone ResNet101, our performance is 0.38\% mIoU higher than Parsing R-CNN even when it changes the backbone to ResNeXt101. %We also show several qualitative results with other works in Fig.~\ref{fig:quality}.
	Our result is 1.6\% mIoU higher than Graphonomy \cite{gong2019graphonomy}, which uses a graph convolution model and adopts a strong backbone DeepLabV3+ \cite{chen2018encoder}. This performance suggests the superiority of our parsing method with the assistance of pose and edge factors, and correlating parsing with pose and edge can introduce contextual cues into human parsing task.
	%收到
	
%	\begin{figure}[t]
%	\begin{center}
%		\centering
%		%\fbox{\rule{0pt}{2in} }
%		\scriptsize
%		\rule{.95\linewidth}{0pt}
%		% 现在篇幅刚好 你先交一版吧 然后再继续改 防止最后来不及
%		\includegraphics[width=0.95\linewidth]{Figures/params_small}
%	\end{center}
%	\vspace{-3mm}
%	\caption{Parameter discussion of $\alpha$ and $\beta$ values in Equation~\ref{eq:loss} on the LIP dataset.}
%	\label{fig:params}
%	\end{figure}

    \begin{table}[tp]
    \centering
        %\fontsize{6.5}{8}\selectfont
    \footnotesize
    %\begin{threeparttable}
    \begin{tabular}{ccccccc}
    \hline
    \multirow{2}{*}{$\alpha$} &
    \multicolumn{6}{c}{$\beta$} \cr  %&\multicolumn{3}{c}{ G}\cr
    \cmidrule(lr){2-7} %\cmidrule(lr){5-7}
    & 0  & 1  & 10 & 50 & 70  & 80 \\ \hline
    0 & 48.72 & 52.08 & 52.77 & 53.10 & 53.98 & 53.59 \\
    1 & 50.98 & 51.15 & 52.03 & 51.54 & 53.78 & 53.13 \\
    2 & 53.08 & 53.52 & 54.08 & 54.01 & \textbf{55.33} & 54.45 \\
    10 & 53.12 & 53.46 & 53.57 & 53.24 & 53.45 & 53.53 \\
    \hline
    \end{tabular}
    \vspace{1mm}
    \caption{Parameter discussion of $\alpha$ and $\beta$ values in Equation~\ref{eq:loss} on the LIP dataset.}
    \vspace{-3mm}
    \label{tab:params}
    %\end{threeparttable}
    \end{table}
	
	\subsection{Evaluation of each component}
	We analyze the parameter sensitivities of our model in Tab.~\ref{tab:params} and validate the effect of each component in Tab.~\ref{tab:perIoU}. 
	
	\textbf{The effect of different loss weights.} The loss values in different branches are crucial to the model. In Tab.~\ref{tab:params}, we test four $\alpha$ values \{0, 1, 2, 10\} with six $\beta$ values \{0, 1, 10, 50, 70, 80\}. $\alpha$ = 0 or $\beta$ = 0 indicates the baseline that removes the edge branch or pose branch from our model. It is observed that adding either edge or pose information to the parsing network brings a significant boost to the baseline. And the model achieves the highest mIoU when $\alpha$ = 2 and $\beta$= 70, which we choose as the final loss weights.
	
	\textbf{The effect of pose and edge cues.} Firstly, we train a baseline model P which only contains parsing branch. In Tab.~\ref{tab:perIoU}, without the contextual cues from pose and edge feature, the baseline model achieves 48.67\% mIoU. We then add an edge/pose branch to the baseline model and concatenate parsing with edge/pose feature to perform prediction, denoted as P+B and P+K. Compared with baseline model P, simple concatenation boosts 2.6\% and 2.12\%  in terms of mIoU, respectively. Particularly, after fusing edge and parsing feature, the performances of some classes which are usually adjacent and have similar appearances (\eg, upper clothes and pants), gain nearly 1.5\% mIoU. These results demonstrate the effectiveness of edge and pose factors to parsing task. And the model P+B+K denotes concatenating both edge and pose feature with parsing feature. It only improves the performance by 0.66\% mIoU compared with P+B, which indicates that even pose and edge factors are necessary for parsing, concatenating all three factors is not an ideal method to sufficiently leverage contextual cues.
	% to accurately localize boundaries and enforce the body part to be consistent with human body structure. 
	%Moreover, , we compare our model with separately concatenating either edge or pose feature with parsing feature, denoted as S+E and S+P. And 

	\begin{figure*}[htb]
	\begin{center}
		\centering
		%\fbox{\rule{0pt}{2in} }
		%\rule{\linewidth}{0pt}
		\includegraphics[width=1\linewidth,height=0.18\textheight]{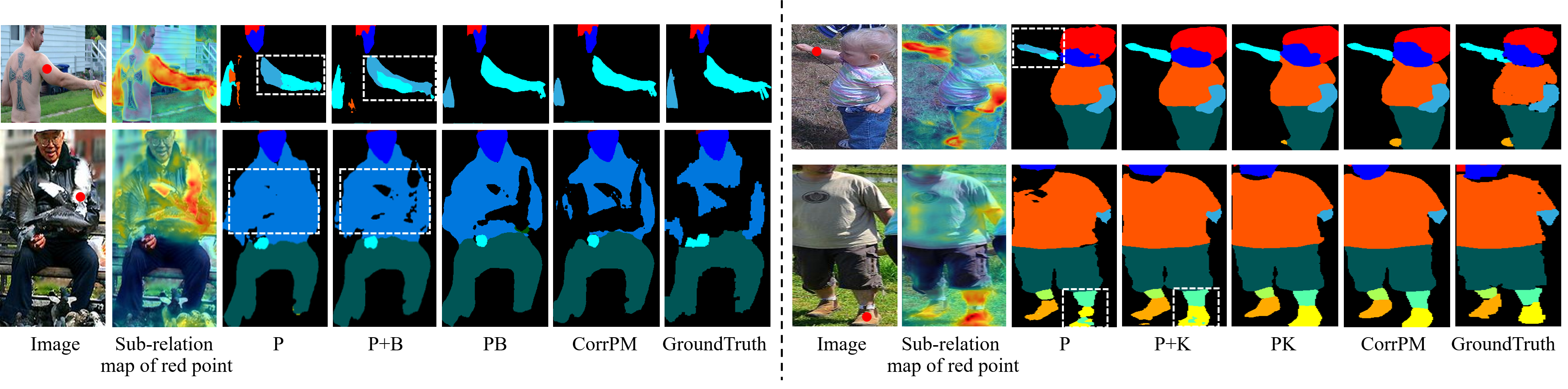}
	\end{center}
	\vspace{-4mm}
	%再交一次，还有10分钟 你交完然后下载你交的 check一下
	\caption{Visualization results of different fusion methods. These images show the benefits of edge/pose information to parsing task. The meaning of symbols is the same as Tab.~\ref{tab:perIoU}. }
	\vspace{-4mm}
	\label{fig:pb}
	\end{figure*}
	%haode，明白，感谢老师！！

	\textbf{The effect of self-correlation.}
	To investigate the effect of the non-local operation, we add a traditional non-local self-attention module at the end of baseline model, denoted as PP. From Tab.~\ref{tab:perIoU}, there is little improvement (0.09\% mIoU) when calculating the relationship within parsing feature itself, and the performance of some classes is reduced such as hat, dress and upper-clothes. It shows that only exploiting the self-correlation of parsing feature is not enough and we need more pivotal factors from pose and edge to boost parsing performance. 

	\begin{figure}[htb]
		\begin{center}
			\centering
			%\fbox{\rule{0pt}{2in} }
			\rule{.9\linewidth}{0pt}
			\includegraphics[width=1\linewidth]{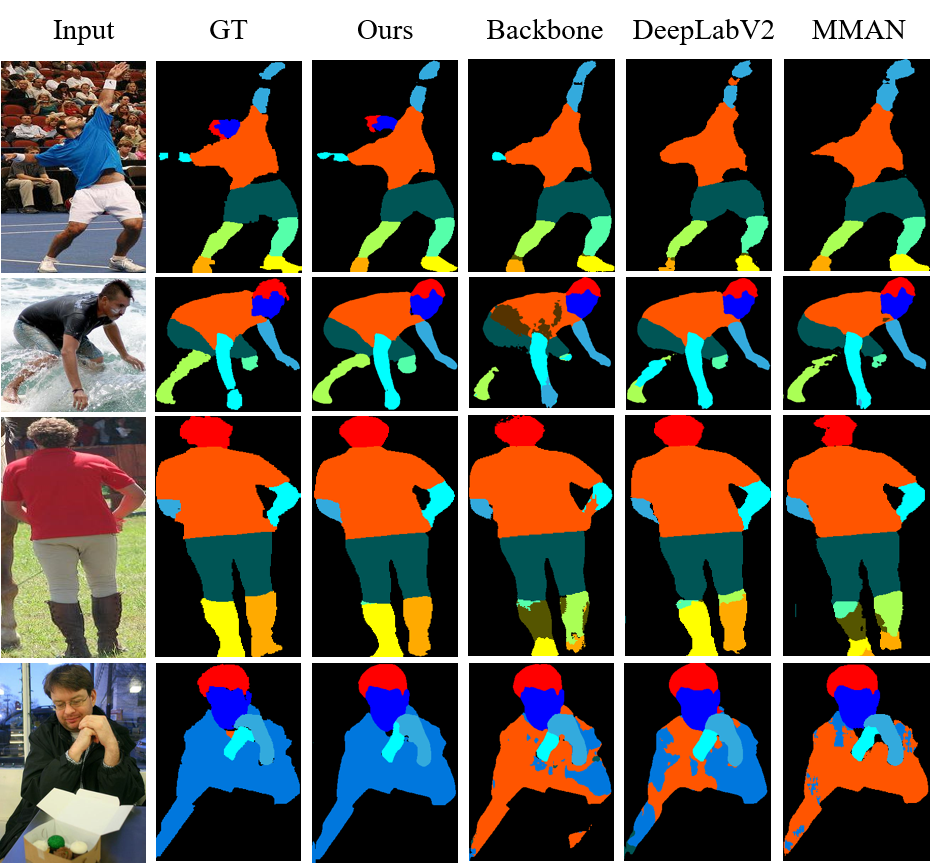}
		\end{center}
		\vspace{-3mm}
		\caption{Visualization of different methods on the LIP dataset. The proposed CorrPM obtains smoother edge prediction and more reasonable body structure results.}
		\label{fig:quality}
	   \vspace{-3mm}
	\end{figure}
	
	\textbf{The effect of correlation among parsing, edge and pose.}
	We conduct two heterogeneous non-local correlations experiments, one is between parsing and edge factor, denoted as PB, and the other is between parsing and pose factor, denoted as PK, to validate the benefits of correlation module to parsing task. The performance improvement is more significant if leveraging the proposed heterogeneous non-local module, yielding 4.44\% and 5.31\% increases in terms of mIoU to baseline model P. And compared with P+B and P+K, the correlation module brings 1.84\% and 3.19\% mIoU gains. And even if we only use either pose or edge features, the result is more than 1.18\% mIoU higher than the concatenation of all the three features, P+B+K.  It is also observed that some categories which are closely related to human body joints are significantly improved by a large margin, which yields about 10\% improvement in terms of mIoU. It shows that our HNL can make sufficient use of edge and pose information to accurately locate the boundary of semantics and maintain the body part geometry. %It is also observed that the correlation operation is more suitable to utilize pose cues. Some categories which are closely related to human body joint are significantly improved by a large margin, about 10\% mIoU. 
	
	\textbf{The effect of integration of multiple tasks.}
	Aggregating the feature maps from multiple tasks will increase the channel number for fusion with parsing feature. Thus experiments are performed to study the efficacy of it. In Tab.~\ref{tab:perIoU}, PBB (PKK) demonstrates the result of fusing two edge (pose) feature maps with parsing feature along channel dimension in HNL. PBB/PKK has the same channel number as CorrPM, while the mIoU is more than 1\% lower than it. %This result illustrates fusing different features from multiple task can boost the performance.
	It shows the improvements are from the integration of multiple tasks rather than the increased channel number. 

	\subsection{Qualitative Results}
	\textbf{The solution of pose and edge to two problems.} 
	As mentioned in Sec.~\ref{section:intro}, there are two problems in human parsing task: inaccurate boundary localization between two adjacent parts and semantics inconsistency of segmented categories. Several images and sub-relation maps are shown in Fig.~\ref{fig:pb} to demonstrate the benefits that the proposed HNL learns from pose and edge information.
	The size of relation map $S$ mentioned in Sec.~\ref{subsection:non_loc} is $HW\times HW$. Hence, for a certain position in the image (marked as red point in Fig~\ref{fig:pb}), the size of its corresponding sub-relation map is $H\times W$. As shown in the left half of Fig~\ref{fig:pb}, some pixels in right arm are wrongly predicted as left arm, while there is no semantic boundary in this region. In the second row, the appearances of the coat and bird are similar so that the baseline model cannot tell them apart. After concatenating edge with parsing feature, the number of error pixels reduces but the boundary is still not clear. When utilizing correlation module, all the semantic edges are rightly predicted. Hence correlating edge with parsing factor can solve inaccurate boundary localization problem.
	From the right part of Fig.~\ref{fig:pb}, the shoes region loses much detail during downsampling process, thus is not correctly classified. Concatenating pose with parsing feature can mitigate this problem. After correlating  with parsing feature, the model obtains the awareness of the position of foot and shoe, hence the shoes classes are segmented correctly. 
	%Furthermore, our HNL model can predict the glass on the face of the slider which is not included in the groundtruth. 
	Therefore, correlating pose with parsing factor can settle the semantics inconsistency matter.

	\textbf{Comparison with the previous methods.}
	We show the quality results in Fig.~\ref{fig:quality} compared with DeepLabV2 \cite{Chen2017deeplab}, MMAN \cite{Luo2018Macro}. Our model outperforms other methods and the predictions are more precise. For example, on the first row, the head and right arms of the person are missing in other methods, while our model correctly predicts them despite the complexity of the background. Besides, with the help of edge information, our framework successfully locates the semantic boundary of the clothes and the legs shown in the second row, and keeps the semantics consistent among upper clothes category. We also observe from the third row that by adding pose information, the model can learn the global body structure of human and accurately identifies the left and right shoes, not legs. Consequently, the proposed HNL effectively employs the relationship of edge, pose and parsing features, and outputs more reasonable and precise results on the human parsing task.
	
	%------------------------------------------------------------------------
	
    \section{Conclusion}
	In this paper, we propose a Correlation Parsing Machine (CorrPM) to take advantage of both semantic edge and human body keypoint features.
	For the two problems in human parsing task, our approach utilizes semantic edge to distinguish the boundary of two adjacent categories and human keypoint to enforce segmented classes to be consistent with body parts.
	With the heterogeneous non-local (HNL) module, the proposed model explores the relationship of edge, pose and parsing factors, and provides the contextual cues for human parsing task.
	The whole model is end-to-end learnable.
	Experiments on three benchmarks demonstrate the effectiveness of the proposed method. Moreover, the proposed system is flexible and easy to be deployed even without pose annotation.
	
	\vspace{0.8mm}
	\textbf{Acknowledgments. }
	%This paper is supported by...
	This work is partially supported by the Beijing Major Science and Technology Project under contract No. Z191100010618003 and National Key Research and Development Program of China under contract No. 2016YFB0402001. We acknowledge Kingsoft Cloud for the helpful discussion and free GPU cloud computing resource support. We are also grateful to Dr Liang Zheng who is the recipient of an Australian Research Council Discovery Early Career Award (DE200101283) funded by the Australian Government.

{\small
\bibliographystyle{ieee_fullname}
\bibliography{egbib}
}

\end{document}